\documentclass{article}

\PassOptionsToPackage{numbers, compress}{natbib}

\usepackage[final]{nips_2017}

\usepackage[utf8]{inputenc} 
\usepackage[T1]{fontenc}    
\usepackage{hyperref}       
\usepackage{url}            
\usepackage{booktabs}       
\usepackage{amsfonts}       
\usepackage{nicefrac}       
\usepackage{microtype}      
\usepackage{amsmath}
\usepackage[linesnumbered,ruled]{algorithm2e}
\usepackage{subcaption}
\usepackage{multirow}
\usepackage{graphicx}
\usepackage{bm}

\graphicspath{ {images/} }

\title{Dynamic Capacity Estimation in Hopfield Networks}

\author{
  Saarthak Sarup, Mingoo Seok\\
  Department of Electrical Engineering\\
  Columbia University\\
  New York, NY 10027 \\
  \texttt{\{ss4754,ms4415\}@columbia.edu} \\  
}

\begin{document}

\maketitle

\begin{abstract}
  Understanding the memory capacity of neural networks remains a challenging problem in implementing artificial intelligence systems. In this paper, we address the notion of capacity with respect to Hopfield networks and propose a \emph{dynamic} approach to monitoring a network's capacity. We define our understanding of capacity as the maximum number of stored patterns which can be retrieved when probed by the stored patterns. Prior work in this area has presented \emph{static} expressions dependent on neuron count $N$, forcing network designers to assume worst-case input characteristics for bias and correlation when setting the capacity of the network. Instead, our model operates simultaneously with the learning Hopfield network and concludes on a capacity estimate based on the patterns which were stored. By continuously updating the crosstalk associated with the stored patterns, our model guards the network from overwriting its memory traces and exceeding its capacity. We simulate our model using artificially generated random patterns, which can be set to a desired bias and correlation, and observe capacity estimates between 93\% and 97\% accurate. As a result, our model doubles the memory efficiency of Hopfield networks in comparison to the static and worst-case capacity estimate while minimizing the risk of lost patterns. 
\end{abstract}

\section{Introduction}

Associative memory storage remains one of the most appealing features of human cognition, enabling complex functions in the hippocampus like efficient memory retrieval and information routing \citep{levy1979synapses, stella2011associative}. Unsurprisingly, associative memory storage has also been an important goal of modern neural networks, and low-latency models like Hopfield networks have traditionally been effective solutions to this problem. 

Described by \citet{little1974existence} and later popularized by \citet{hopfield1982neural}, (Little-)Hopfield networks are a form of recurrent neural networks which consist of $N$ fully-connected McCulloch-Pitts neurons. In addition to this topology, Hopfield networks are typically designed with the constraint that no neuron has a connection with itself and that all connections are symmetric. They have the ability to store binary patterns of the form $\{+1,-1\}^N$, and are commonly learned according to the standard Hebb rule, formalized as a sum-over-outer-products. These patterns are then recalled by probing the network with a possibly noisy version of the desired pattern and allowing the network to relax into a low energy state according to its attractor dynamics \citep{aiyer1990theoretical, nasrabadi1992hopfield, storkey1999basins}.

A common question asked of these networks is their capacity --- how many patterns can a $N$-neuron Hopfield network reliably store? Increasing the number of learned patterns introduces additional terms called crosstalk \citep{macgregor1991cross}, which can be intuitively understood as the influence of stored patterns on the network response to a desired pattern. The accumulation of these crosstalk terms results in higher order instabilities in the stored patterns, as they produce spurious states as undesirable fixed points in the $N$-dimensional attractor space \citep{wu2012storage}. We explore the mechanics of this behavior in Section \ref{crosstalk}, which enables our effective capacity estimation for the Hopfield network. 

Here we define this term, \emph{capacity}, as the maximum number of stored patterns $M$ which can be perfectly recalled when probed with the desired pattern. We consider alternate definitions and their relation to ours in Section \ref{related}. Nevertheless,  mathematically rigorous analyses have been conducted on this question using a compatible definition of capacity, producing coarse estimates like $M \approx 0.14N$ to more precise expressions like $M = \frac{N}{4\log N}$ \citep{amit1985storing,mceliece1987capacity}. Unfortunately, these static estimates fair poorly on a number of relevant metrics like accuracy and efficiency, as we observe in Section \ref{experiment}.

\begin{figure}[t]
    \centering
    \includegraphics[scale=0.14]{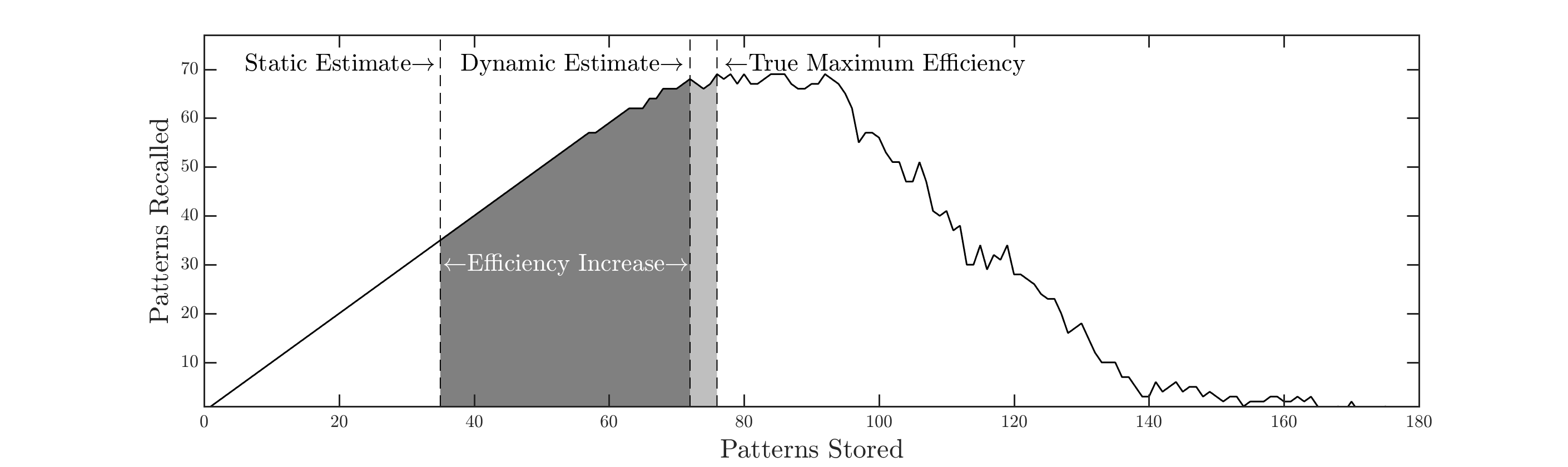}
    \caption{Example capacity simulation for $N=1000$ Hopfield network. The dark shaded region indicates the efficiency improvement using our proposed dynamic model in comparison to a static estimate with the worst-case correlation and bias. Lightly shaded region shows the minimal efficiency difference between our dynamic model and the true maximum efficiency.}
    \label{intro_image}
\end{figure}

Later analyses by L{\"o}we have extended prior work by incorporating the possibility for bias \citep{lowe1999storage} or correlation \citep{lowe1998storage} in the stored patterns. Their theoretical results propose a reduction in memory capacity as either bias or correlation diverge from the ideal i.i.d. random pattern case, and experimental simulations testing these hypotheses support their claims \citep{wilson1998effects}.

Since these static capacity models do little in the way of accounting for naturally varying patterns, Hopfield networks using a standard Hebb learning rule are rarely effective in practice. Designers hoping to perfectly recall all stored patterns are then forced to drastically underestimate the capacity for the worst-case scenario, as we depict in Figure \ref{intro_image} above. This is of particular importance in VLSI design, as hardware designers intending to use Hopfield networks as a form of content-addressable memory use an impractically large amount of memory to store the network's weight matrix relative to the network's information capacity \citep{draghici2000neural, theocharides2004embedded, verleysen1989neural}.

In place of the standard static estimates for memory capacity defined in Section \ref{related}, we present a novel dynamic technique for monitoring the capacity of a Hopfield network in real time. Our estimation model is a function of both the current state of the network, as represented by the network's weight matrix, and the incident pattern to be stored. By testing the current network's response to a new pattern, we calculate a fraction of the total crosstalk in our system, allowing us to observe the network's distance to its capacity. This behavior is more quantitatively analyzed in Section \ref{approach}. We then compare our results on the same metrics used to analyze static estimates and note an increase from an average of 53\% accuracy to an average of 94\% accuracy using our dynamic capacity model. Moreover, we observe a doubling of memory efficiency when compared to the static models that assume worst-case capacity estimates. 

\section{Related work}
\label{related}
Hopfield and other early researchers like Amit, Gutfreund and Sompolinsky considered a notion of capacity defined for an $N$-neuron Hopfield network by a critical value $\alpha_c$, such that for $M \leq \alpha_cN$, nearly all $M$ patterns are memorized, and for $M > \alpha_cN$, nearly all $M$ patterns are forgotten \citep{hopfield1982neural}. By comparing Hopfield networks to physical spin-glass models and assuming i.i.d. random patterns, they conclude with a calculation of $\alpha_c \approx 0.138$ and therefore $M = 0.138N$ represents the capacity of a Hopfield network \citep{amit1985storing}. We find this definition of capacity too coarse for ensuring minimal memory loss and thus is not considered any further in this work.

The question of capacity was brought up again by McEliece, Posner, Rodemich and Venkatesh with the more precise definition that capacity is the maximum number of i.i.d. random patterns $M$ which can be stored and perfectly recalled. They extend their analysis to considering initial probes with patterns at a distance of $d = \delta N$ Hamming distance away from one of the $M$ patterns, where $0\leq d < N/2$. Their exact derivation is not useful here but can be found in \citep{mceliece1987capacity}. Their final expression takes the following form:
\begin{equation}M = \frac{(1-2\delta)^2}{4}\frac{N}{\log N}\end{equation}

L{\"o}we broke from the assumption of i.i.d. random patterns to consider the effect of pattern bias \citep{lowe1999storage} or correlation \citep{lowe1998storage} on the capacity. He defined a bias $b$ for the $i$-th value of the $\mu$-th pattern by $\mathbb{P}(x_i^\mu = 1) = b$. He modeled both semantic and spatial correlations as homogeneous Markov chains, defining the correlation $c$ as $\mathbb{P}(x_i^\mu = x_i^{\mu -1}) = c$. For both cases, biased and correlated, L{\"o}we expresses the capacity as $M = \nicefrac{N}{\gamma \log N}$, where $\gamma$ is defined based on the bias or correlation of the stored patterns. 
\begin{equation}\gamma > \frac{3}{8b^2(1-b)^2}\end{equation}
\begin{equation}\gamma > 48c(1-c)\big(c^2 + (1-c)^2\big)\end{equation}

The definition of capacity employed in this paper is the maximum number of stored patterns that a Hopfield network can retrieve when probed with the desired pattern. This definition is compatible with the definitions used by McEliece et. al. and by L{\"o}we, and are used as baselines to which we compare our dynamic estimation model in Section \ref{experiment}. For the static estimate by McEliece et. al., our definitions are equivalent to the case where $\delta = 0$ and thus we use the expression $M = \frac{N}{4\log N}$. For the estimate by L{\"o}we, we recognize his lower bound for the parameter $\gamma$ and thus we use both its lower bound for the appropriate value of $b$ or $c$.

\section{Random pattern generation}
\label{rpg}
Since we examine the behavior of Hopfield networks on artificial data, it is important to precisely define how we generate our test patterns. Each of the $M$ patterns is represented by an $N$-dimensional column vector $x^\mu = [x_1^\mu \ \ x_2^\mu \ \ ... \ \ x_N^\mu]^T$, where each element $x_i^\mu$ is chosen from the set $\{+1, -1\}$. In the i.i.d. case, each element $x_i^\mu$ is chosen independently of all other elements for $1 \leq i \leq N$ and $1 \leq \mu \leq M$, and with equal probability for either $+1$ or $-1$.

In order to produce data with bias and/or correlation, we rely on techniques introduced by L{\"o}we \citep{lowe1999storage}. We introduce bias by setting the probability that a particular element is either $+1$ or $-1$, formalized as $\mathbb{P}(x_i^\mu = +1) = b$ and thus $\mathbb{P}(x_i^\mu = -1) = 1-b$. In order to introduce correlation, we define a parameter $c \in [0,1]$ and consider semantic correlations between successive patterns as Markov chains. We then combine these two parameters based on a method described by \citet{clopath2012storage}. Explicitly, we define the probabilities as follows:
\begin{align}
    \mathbb{P}(x_i^\mu = +1 \vert \ x_i^{\mu-1} = +1) &= b + c(1-b) \\
    \mathbb{P}(x_i^\mu = +1 \vert \ x_i^{\mu-1} = -1) &= (1-c)b \\
    \mathbb{P}(x_i^\mu = -1 \vert \ x_i^{\mu-1} = +1) &= (1-c)(1-b) \\
    \mathbb{P}(x_i^\mu = -1 \vert \ x_i^{\mu-1} = -1) &= 1-b + cb
\end{align}

This approach allows us to consider the input patterns taking the form of $N$ independent two-state Markov chains. We can then reason about the correlation between successive patterns and quantify the crosstalk in the Hopfield network as a function of the parameters $b$ and $c$. 
\subsection{Pattern statistics}
Each two-state chain can be represented by a $2\times2$ transition matrix $P$ where elements $p_{ij}$ contain the probability of moving from state $i$ to state $j$ in one step.
\begin{equation}
    P = 
    \begin{bmatrix}
    1-\alpha & \alpha \\
    \beta    & 1-\beta
    \end{bmatrix}
\end{equation}
\begin{align}
    \alpha &= \mathbb{P}(x_i^\mu = -1 \vert \ x_i^{\mu-1} = +1) = (1-c)(1-b) \\
    \beta  &= \mathbb{P}(x_i^\mu = +1 \vert \ x_i^{\mu-1} = -1) = (1-c)b
\end{align}
We can extend this transition matrix to $n$ steps as $P^{(n)}$. Recognizing the eigenvalues of $P$ as 1 and $1-\alpha-\beta$, we can calculate $P^{(n)}$
\begin{equation}
    P^{(n)} = 
    \begin{bmatrix}
    \frac{\beta}{\alpha+\beta} + \frac{\alpha}{\alpha+\beta}(1-\alpha-\beta)^n & \frac{\alpha}{\alpha+\beta} - \frac{\alpha}{\alpha+\beta}(1-\alpha-\beta)^n \\
    \frac{\beta}{\alpha+\beta} - \frac{\beta}{\alpha+\beta}(1-\alpha-\beta)^n & \frac{\alpha}{\alpha+\beta} + \frac{\beta}{\alpha+\beta}(1-\alpha-\beta)^n
    \end{bmatrix}
\end{equation}

A useful quantity in our later analysis of crosstalk in Section \ref{crosstalk} is the correlation between two stored patterns separated by $n$ steps, quantified as $\sum_i^N x_i^\mu x_i^{\mu-n}$. As a result, we are interested in describing $\mathbb{P}(x_i^\mu x_i^{\mu-n} = 1)$ as well as the expected value $\mathbb{E}(x_i^\mu x_i^{\mu-n})$. Writing $\mathbb{P}(x_i^\mu x_i^{\mu-n} = 1)$ as a function of $n$, $\rho(n)$, allows us to simplify the expression for the expected value to $\mathbb{E}(x_i^\mu x_i^{\mu-n}) = e(n) = 2\rho(n) -1$. We calculate this function $\rho(n)$ as follows:
\begin{equation}
    \rho(n) = bP_{11}^{(n)} + (1-b)P_{22}^{(n)}
\end{equation}

We can gain a more intuitive understanding of these equations by evaluating their order for increasing values of $n$. This corresponds to thinking about the correlation between two patterns separated by $n$ units of time. Since these correlations are not dependent on our bias $b$, we leave $b$ a constant and analyze the behavior as a function of our correlation $c$.
\begin{align*}
    \rho(1) = bP_{11}^{(1)} + (1-b)P_{22}^{(1)} = \mathcal{O}(c) \quad   &\Rightarrow \quad e(1) = \mathcal{O}(c) \\
    \rho(2) = bP_{11}^{(2)} + (1-b)P_{22}^{(2)} = \mathcal{O}(c^2)
    \quad &\Rightarrow \quad e(2) = \mathcal{O}(c^2) \\
    & \ \ \vdots \\
    \rho(n) = bP_{11}^{(n)} + (1-b)P_{22}^{(n)} = \mathcal{O}(c^n)
    \quad &\Rightarrow \quad e(n) = \mathcal{O}(c^n)
\end{align*}

It becomes evident then that for $0 < c < 1$, the expected value of the term $x_i^\mu x_i^{\mu-n}$ decreases for larger distances $n$. This behavior fits our understanding of the Markov property where the future state is dependent on the current state, and thus for increasing separations in time the successive patterns grow decreasingly correlated.

\section{Dynamic capacity estimation model}
As we will analyze in this section, the fundamental source of memory loss in Hopfield networks arises from an accumulation of crosstalk. Since crosstalk is dependent on the patterns being stored, which can vary drastically in bias and correlation, traditional static approaches to estimating capacity are often inaccurate at predicting the capacity of the network for any particular trial. Taking a dynamic approach to measuring the crosstalk in real time enables us to adapt our estimate of the network's capacity to the patterns being stored and halt its operation before stored memories are lost.

\subsection{Quantitative analysis of crosstalk}
\label{crosstalk}
Consider an $N$-neuron Hopfield network with $N^2$ synapses stored in an $N\times N$ weight matrix $W$. We plan to store $M$ patterns of length $N$ as $x^\mu$ for $1 \leq \mu \leq M$, and represent the network after storing the $\mu$-th pattern as $W^{\mu}$. Initially, $W_{ij}^{(0)} = 0$. After the first pattern $x^{(1)}$ is stored, we can write the weight matrix as $W_{ij}^{(1)} = \frac{1}{N}x_i^{(1)}x_j^{(1)}$, and after $\mu$ patterns, we write the matrix as:
\begin{equation}
    W_{ij}^{\mu} = \frac{1}{N}\sum_{\mu = 1}^{M}x_i^{\mu}x_j^{\mu}
\end{equation}

Suppose we now attempted to recall all $M$ patterns. We calculate their activation $a^{\mu} = \sum_{j\neq i}^N W_{ij}x_j^{\mu}$ and note that, at least to first order recurrence in the network, $\hat{x}^{\mu} = \text{sgn}(a^{\mu})$.
\begin{align*}
    a_i^{(1)} &= \sum_{j \neq i}^{N}W_{ij}^{(M)}x_j^{(1)} = x_i^{(1)} + \frac{x_i^{(2)}}{N}\sum_{j \neq i}^{N}x_j^{(2)}x_j^{(1)} +  ... + \frac{x_i^{(M)}}{N}\sum_{j \neq i}^{N}x_j^{(M)}x_j^{(1)} \\
    a_i^{(2)} &= \sum_{j \neq i}^{N}W_{ij}^{(M)}x_j^{(2)} = \frac{x_i^{(1)}}{N}\sum_{j \neq i}^{N}x_j^{(1)}x_j^{(2)} + x_i^{(2)} +  ... + \frac{x_i^{(M)}}{N}\sum_{j \neq i}^{N}x_j^{(M)}x_j^{(2)} \\
    & \ \ \vdots \\
    a_i^{(M)} &= \sum_{j \neq i}^{N}W_{ij}^{(M)}x_j^{(M)} = \frac{x_i^{(1)}}{N}\sum_{j \neq i}^{N}x_j^{(1)}x_j^{(M)} +
    \frac{x_i^{(2)}}{N}\sum_{j \neq i}^{N}x_j^{(2)}x_j^{(M)} +  ... + x_i^{(M)}
\end{align*}
We notice that along with the desired component $x_{i}^\mu$ we also recover crosstalk terms which correspond to the contribution of other memories on $x_{i}^\mu$. We also notice that the crosstalk varies based on the neuron $i$, and so we define $\kappa_n^m$ corresponding to the total crosstalk affecting the $m$-th pattern at neuron $n$.
\begin{equation}
    \kappa_{n}^m = \frac{1}{N}\sum_{\mu \neq m}^{M}\sum_{j\neq n}^{N}x_n^{\mu}x_j^{\mu}x_j^{m}
\end{equation}

Modeling our input data as a Markov chain allowed us to express the expected value of the term $x_j^\mu x_j^m$ as $e(\mu-m)$. We can treat $\kappa_n^m$ as a random variable and express its expected value as
\begin{equation}
    \mathbb{E}(\kappa_n^m) = \frac{N-1}{N}\sum_{\mu\neq m}^{M}x_n^\mu \ e(\mu-m)
\end{equation}

Knowing that the function $e(n)$ decreases for increasing values of $n$ based on the correlation of the patterns allows us to use future measurements of $\kappa_n^m$ and estimate the distribution of crosstalk across all of its terms, as we will show in Section \ref{approach}.

Another notable feature of crosstalk is that it comes in two variants: constructive and destructive. Since the recovered bit $\hat{x}_i^\mu$ depends on the sign of $a_i^\mu$, it is possible that $\kappa_i^\mu$ is the same sign as the desired bit, and thus we get that $\hat{x}_i^\mu = \text{sgn}(a_i^\mu) = x_i^\mu$. Thus we are only concerned with the case when $-x_i^\mu\kappa_i^\mu > 1$. When this occurs, the $i$-th neuron in the network returns the incorrect bit for the pattern $x^\mu$ and causes the overall network to deviate from its trajectory towards the desired fixed point in $\mathbb{R}^N$ and instead settle in an incorrect low energy state.

\subsection{Our approach}
\label{approach}
In order to monitor a Hopfield network's capacity, suppose we maintain a $N\times M$ matrix $\chi$ which contains the destructive crosstalk for the Hopfield network, where we define the elements as $[\chi]_{nm} = -x_n^m\kappa_n^m$. Consider the result of storing an $M+1$ pattern as $W_{ij}^{(M+1)} = W_{ij}^{(M)} + \frac{1}{N} x_i^{(M+1)}x_j^{(M+1)}$. When attempting to recover this new pattern through the networks activation to $x^{(M+1)}$, we calculate $a_i^{(M+1)} = \sum_{j\neq i}^{N}W_{ij}^{(M)}x_j^{(M+1)} + x_i^{(M+1)}$. Using this notation, we see the crosstalk for the $M+1$ pattern in all $i$ neurons already bundled together as $\kappa_i^{M+1} = \sum_{j\neq i}^{N}W_{ij}^{(M)}x_j^{(M+1)}$. Thus we can use the following iterative approach to calculate exactly the destructive crosstalk associated with each new pattern $x^{(M+1)}$, appending a column to our matrix $\chi$ in order to keep track of the new crosstalk:
\begin{equation}
    [\chi]_{i,M+1} = -x_i^{(M+1)}\sum_{j\neq i}^{N}W_{ij}^{(M)}x_j^{(M+1)}
\end{equation}

However, this calculation alone is not sufficient for monitoring the crosstalk for all patterns, as adding a pattern to the network also introduces crosstalk for the previously stored patterns. After storing the $M+1$ pattern, we must update the crosstalk for all $M$ patterns at all $N$ neurons with the following quantities.
\begin{align*}
    [\chi']_{i,1} &= [\chi]_{i,1} - \frac{x_i^{(1)}x_i^{(M+1)}}{N}\sum_{j\neq i}^{N}x_j^{(M+1)}x_j^{(1)} \\
    [\chi']_{i,2} &= [\chi]_{i,2} - \frac{x_i^{(2)}x_i^{(M+1)}}{N}\sum_{j\neq i}^{N}x_j^{(M+1)}x_j^{(2)} \\
    &\ \ \vdots \\
    [\chi']_{i,M} &= [\chi]_{i,M} - \frac{x_i^{(M)}x_i^{(M+1)}}{N}\sum_{j\neq i}^{N}x_j^{(M+1)}x_j^{(M)}
\end{align*}
We can note that the sum of these crosstalk terms becomes $-\frac{x_i^{(M+1)}}{N}\sum_{\mu = 1}^{M}\sum_{j\neq i}^{N}x_i^{\mu}x_j^{(M+1)}x_j^{\mu} = -\frac{x_i^{(M+1)}}{N}\sum_{\mu = 1}^{M}\sum_{j\neq i}^{N}x_i^{\mu}x_j^{\mu}x_j^{(M+1)} = -x_i^{(M+1)}\sum_{j\neq i}^{N}W_{ij}^{(M)}x_j^{(M+1)}$. Thus the new quantity we have calculated as the crosstalk for pattern $M+1$ is also the total crosstalk to be distributed across the other $M$ patterns. 

Using our knowledge of the pattern's bias $b$ and correlation $c$, we know the expected distribution of $[\chi]_{i,\mu}$ for all $1\leq \mu \leq M$ from the calculated quantity $[\chi]_{i,M+1}$. We know that our function $e(n)$, which describes the expected correlation between two patterns separated by $n$ units in time, scales like $c^n$. As a result, we can understand the individual crosstalk terms as follows, though of course the final expressions which our model uses are functions of $\alpha$ and $\beta$.
\begin{align*}
    \Big([\chi']_{i,1} - [\chi]_{i,1}\Big) &\propto c^M[\chi]_{i,M+1} \\
    \Big([\chi']_{i,2} - [\chi]_{i,2}\Big) &\propto c^{M-1}[\chi]_{i,M+1} \\
    & \ \ \vdots \\
    \Big([\chi']_{i,M} - [\chi]_{i,M}\Big) &\propto c[\chi]_{i,M+1}
\end{align*}

\begin{figure}[h]
    \centering
    \begin{subfigure}{0.3\textwidth}
        \centering
        \includegraphics[scale=0.17]{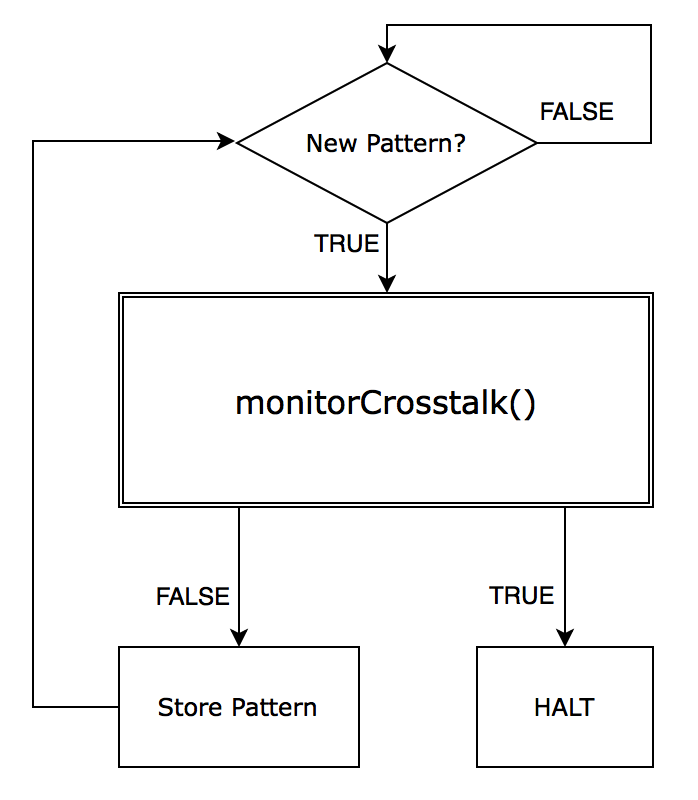}
        \caption{}
        \label{diagram}
    \end{subfigure}
    \quad \quad \quad
    \begin{subfigure}{0.5\textwidth}
        \centering
        \setlength{\interspacetitleruled}{-.4pt}
        \begin{algorithm}[H]
        \SetKwInOut{Output}{Output}

        \textbf{function} \underline{monitorCrosstalk}$(W,x,m, \chi)$\;
        \Output{Boolean for under/over capacity}
        \vspace*{.1cm}
        \eIf{$m$ equals 0}
          {
            $\chi \leftarrow x * Wx$ 
          }
          {
            $a \leftarrow x * Wx$\\
            Calculate row-vector $\rho(n) \ \forall n \ | \ 1 \leq n \leq m$ \\
            $\chi \leftarrow \chi + a \otimes \rho$ \\
            $\chi \leftarrow [\chi \ a]$
          }
        \Return{any$(\chi \geq 1)$}

        \end{algorithm}
        \caption{}
        \label{algorithm}
    \end{subfigure}
    \caption{High level diagram in Figure \ref{diagram} describes how our model interacts with the learning Hopfield network. Figure \ref{algorithm} recounts the underlying pseudocode used by our model to update the crosstalk and monitor the network's capacity.}
\end{figure}

Updating our destructive crosstalk matrix $\chi$ in this way, written explicitly in Figure \ref{algorithm}, allows us to iteratively and continuously update the crosstalk for each memory at each neuron. Checking if the network is at capacity thus becomes a simple question of checking if $[\chi]_{nm} \geq 1$ for any $1 \leq n \leq N$ and $1\leq m\leq M+1$, and allows us operate alongside a Hopfield network as in Figure \ref{diagram} and halt the network before it has exceeded its capacity.

\section{Experiment}
\label{experiment}

\subsection{Correctness of our random pattern generator}
To verify our pattern generator, we consider bias as $b \in [0.5, 1)$ and correlation as $c \in [0, 1)$. We measure the bias of the generated patterns from the mean of each produced pattern, and the correlation from the mean covariance between successive patterns. We note unexpected deviations for highly biased and highly correlated patterns, and thus do not test our model in this regime.

\begin{figure}[h]
    \centering
    \begin{subfigure}{0.4\textwidth}
        \includegraphics[width=0.81\linewidth, height=3.5cm]{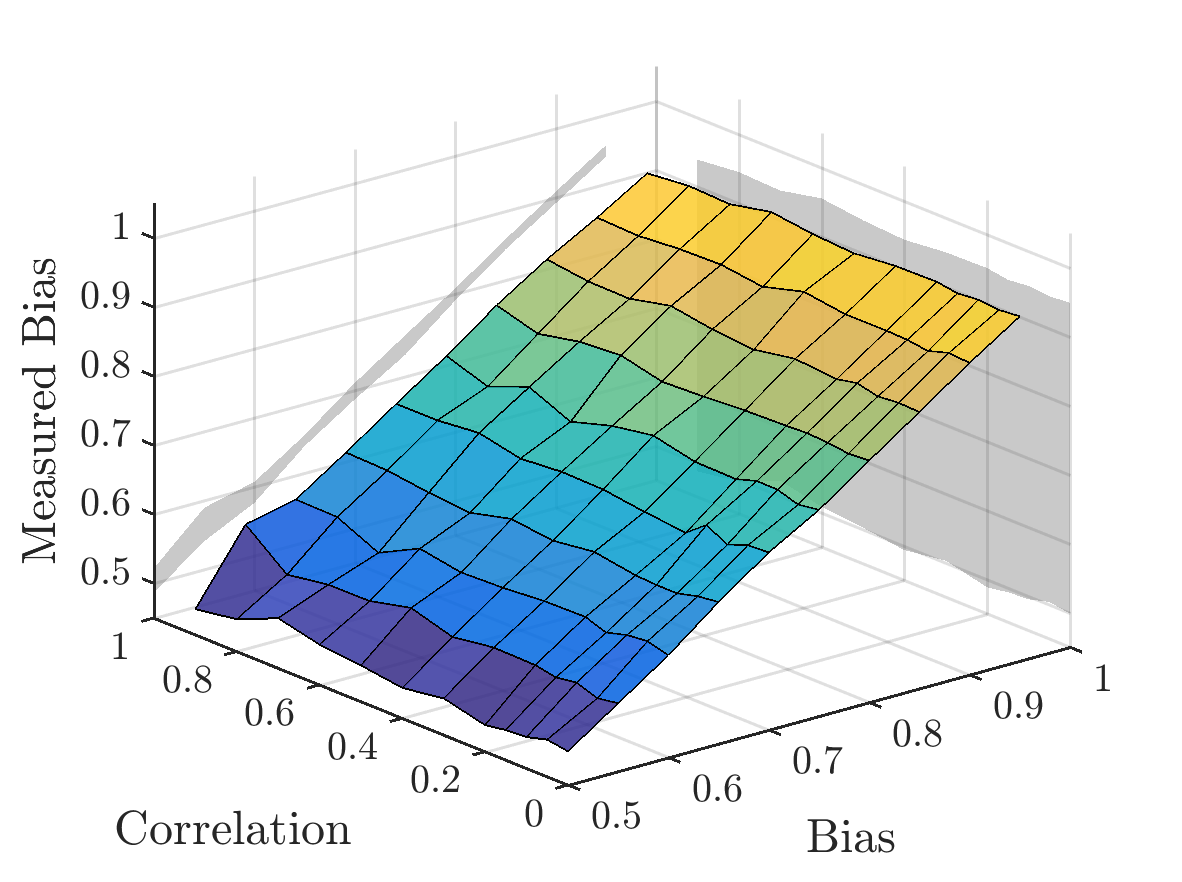}
        \caption{}
        \label{bias}
    \end{subfigure}
    \quad
    \begin{subfigure}{0.4\textwidth}
        \includegraphics[width=0.86\linewidth, height=3.5cm]{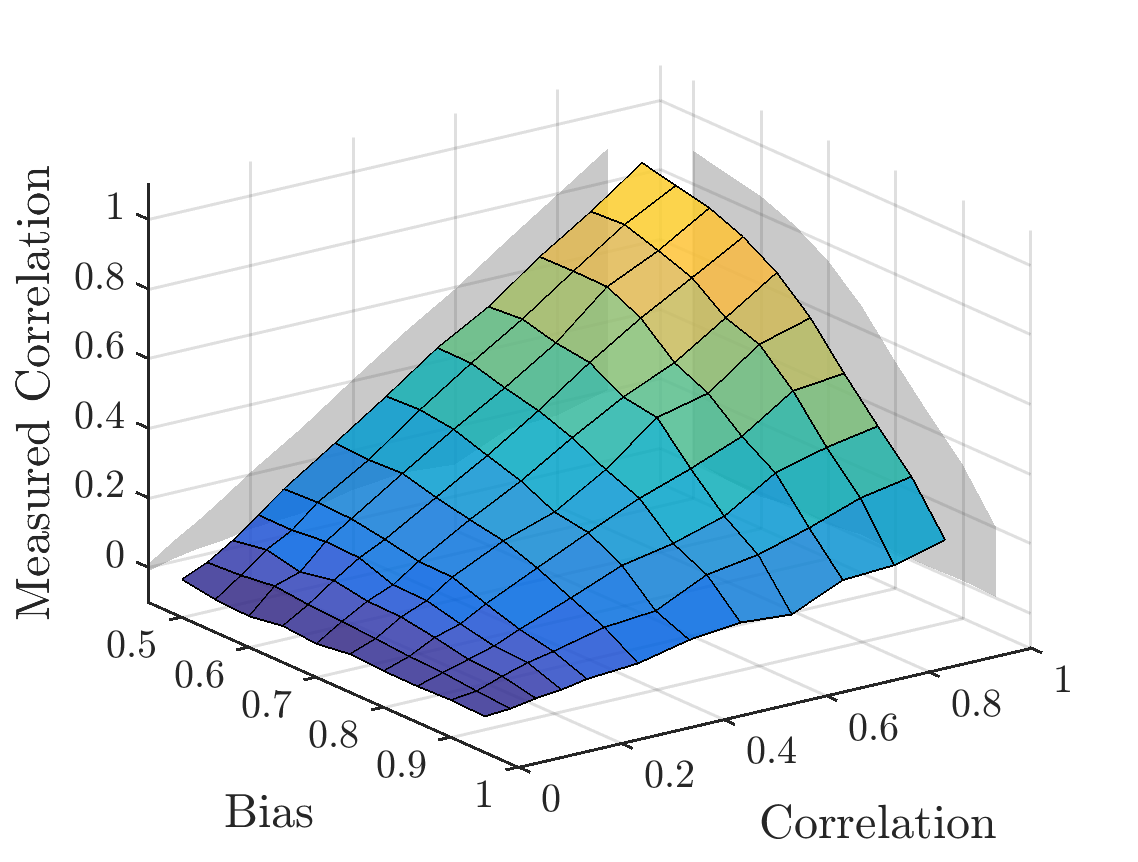}
        \caption{}
        \label{corr}
    \end{subfigure}
    \caption{Measured bias in Figure \ref{bias} and measured correlation in Figure \ref{corr}, shown only for patterns with length $N=500$ to avoid redundancy.}
\end{figure}
\subsection{Accuracy on random unbiased, uncorrelated data}
\label{accuracy}

To simulate the behavior of our model, we fix our values for network size $N$, pattern bias $b$, and correlation $c$. We then generate random patterns according to the desired bias and correlation, and begin the process of simulating our estimator. Each incident pattern is fed through our estimator and stored in the network as is outlined in Section \ref{approach}. We complete each trial once the estimator flags the capacity and the network crosses its true capacity, allowing us to compare our model's estimate.

We first look to analyze the behavior of our model on the ideal case of unbiased and uncorrelated data, using $b = 0.5$ and $c = 0$. This case is also applicable for the case when the biased pattern case when we use the zero-centering learning rule. We run 20 trials at neuron counts of $N = 500$, $N = 1,000$, $N=3,000$, $N=5,000$, and $N=10,000$, corresponding to network sizes ranging from $N^2 = 25\times10^4$ to $N^2 = 10^8$ synapses. We quantify the accuracy of our estimate as the ratio of our estimate $\hat{C}$ to the true capacity $C_0$, and compare our results on this metric to those from the static model proposed by McEliece et. al. as $\hat{C} = \nicefrac{N}{4\log N}$. 

\begin{table}[h]
  \caption{Capacity simulation results with for unbiased, uncorrelated patterns}
  \label{main-table}
  \centering
  \begin{tabular}{ccc}
    \toprule
    Size                          & Model                   & Accuracy         \\
    \midrule
    \multirow{2}{*}{$N = 1,000$}   & $\nicefrac{N}{4\log N}$ & $52\% \pm 1.9\%$  \\
                                  & Dynamic                 & $\mathbf{97\% \boldsymbol{\pm} 5.7\%}$ \\
    \midrule
    \multirow{2}{*}{$N = 3,000$}   & $\nicefrac{N}{4\log N}$ & $53\% \pm 0.8\%$  \\
                                  & Dynamic                 & $\mathbf{93\% \boldsymbol{\pm} 6.1\%}$ \\
    \midrule
    \multirow{2}{*}{$N = 5,000$}   & $\nicefrac{N}{4\log N}$ & $54\% \pm 0.6\%$\\
                                  & Dynamic                 & $\mathbf{94\% \boldsymbol{\pm} 4.3\%}$ \\
    \midrule
    \multirow{2}{*}{$N = 10,000$}  & $\nicefrac{N}{4\log N}$  & $54\% \pm 0.6\%$\\
                                  & Dynamic                  & $\mathbf{95\% \boldsymbol{\pm} 4.9\%}$ \\
    \bottomrule
  \end{tabular}
\end{table}

\paragraph{Results} We can immediately recognize the approximately $40\%$ improvement in capacity estimate from the static model to our proposed dynamic model in Table \ref{main-table} above. Interestingly, the model proposed by McEliece et. al. does not seem to degrade for larger values of $N$, suggesting the general expression is likely a good approximation for a Hopfield network's capacity. Nevertheless, tuning this expression for different values of $N$ would not take into account natural variations in bias and correlation, and thus is not a practical solution for accurately predicting the network's capacity. Instead, our dynamic model showed the most promising results, maintaining an accuracy between about $95\%$ across all tested network sizes. 

\subsection{Efficiency on biased or correlated data}
\label{efficiency}

Allowing for variance in the bias or correlation of our stored patterns introduces a notion of the worst-case capacity estimate. To quantify the reduction in memory efficiency when utilizing the worst-case capacity estimate, we define efficiency as the ratio between the estimated capacity $\hat{C}$ to the number of synapses $N^2$. 
\begin{equation}
\eta = \frac{\hat{C}}{N^2}
\end{equation}

This metric is analogous to the ratio of the amount of information stored to the number of bits needed to store the information, and the worst-case capacity estimate where $\hat{C} < C_0$ corresponds to a reduction in efficiency.

We test our dynamic model in several non-ideal regimes corresponding variable levels of bias and correlation. Specifically, we test our model on values of $\bar{b} = \{0.5, 0.55, 0.6, 0.65\}$ and $\sigma_b = \{0.03, 0.05\}$, and on $\bar{c} = \{0.05, 0.1, 0.15, 0.2, 0.25\}$ and $\sigma_c = \{0.03, 0.05\}$. We compare our efficiency both to the maximum efficiency, $\nicefrac{C_0}{N^2}$, and to L{\"o}we's static capacity model $\hat{C} = \nicefrac{N}{\gamma\log N}$. We choose $\gamma$ for the static model based on the worst case value of bias or correlation, $b = \bar{b}+3\sigma_b$ and $c = \bar{c}+3\sigma_c$. For brevity we report a sampling of these tests for a network size of $N^2 = 9\times10^6$, though they are representative across our previously examined range of $N$. For patterns with high bias and correlation, we are unable to faithfully compare our model's efficiency to the static model's efficiency because the static model vastly overestimates the memory capacity, leading to a rapid destruction of all previously stored patterns and an effective efficiency of 0\%.

\paragraph{Results} It is clear from our results in Table \ref{second-table} that our dynamic model has a higher efficiency than the static model. Moreover, our model's efficiency is within one standard deviation of the maximum efficiency, which fully utilizes the capacity of the Hopfield network. More precisely, we report an average of two-fold improvement on the efficiency as compared to the static model.

\begin{table}[h]
    \caption{Capacity simulation results in $N=3000$ Hopfield network for biased or correlated patterns}
    \label{second-table}
    \centering
    \begin{tabular}{ccccc}
        \toprule
         $\bar{b}\pm \sigma_b$          & $\bar{c}\pm \sigma_c$           & Model                         & Efficiency $(\times 10^{-5})$        & Maximum Efficiency $(\times 10^{-5})$\\
         \midrule
         \multirow{2}{*}{$0.5\pm0.03$}  & \multirow{2}{*}{$0\pm0$}        & $\nicefrac{N}{\gamma\log N}$  & $0.595$         & \multirow{2}{*}{$1.86\pm0.041$} \\
                                        &                                 & Dynamic                       & $\mathbf{1.75 \boldsymbol{\pm} 0.141}$ &  \\
         \midrule
         \multirow{2}{*}{$0.55\pm0.05$} & \multirow{2}{*}{$0\pm0$}        & $\nicefrac{N}{\gamma\log N}$  & $0.378$         & \multirow{2}{*}{$0.66\pm0.085$} \\
                                        &                                 & Dynamic                       & $\mathbf{0.645 \boldsymbol{\pm} 0.086}$ &  \\
         \midrule
         \multirow{2}{*}{$0.65\pm0.03$} & \multirow{2}{*}{$0\pm0$}       & $\nicefrac{N}{\gamma\log N}$  & $-$         & \multirow{2}{*}{$0.11\pm0.002$} \\
                                    &                                     & Dynamic                       & $\mathbf{0.10             \boldsymbol{\pm} 0.009}$ &  \\
         \midrule
         \multirow{2}{*}{$0.5\pm0$}     & \multirow{2}{*}{$0.05\pm0.03$}  & $\nicefrac{N}{\gamma\log N}$  & $0.833$         & \multirow{2}{*}{$1.77\pm0.046$} \\
                                        &                                 & Dynamic                       & $\mathbf{1.70 \boldsymbol{\pm} 0.106}$ &  \\
         \midrule
         \multirow{2}{*}{$0.5\pm0$} & \multirow{2}{*}{$0.1\pm0.05$}       & $\nicefrac{N}{\gamma\log N}$  & $0.694$         & \multirow{2}{*}{$1.42\pm0.040$} \\
                                    &                                     & Dynamic                       & $\mathbf{1.41             \boldsymbol{\pm} 0.117}$ &  \\
         \midrule
         \multirow{2}{*}{$0.5\pm0$} & \multirow{2}{*}{$0.25\pm0.05$}       & $\nicefrac{N}{\gamma\log N}$  & $-$         & \multirow{2}{*}{$0.44\pm0.060$} \\
                                    &                                     & Dynamic                       & $\mathbf{0.42             \boldsymbol{\pm} 0.093}$ &  \\
         \bottomrule
    \end{tabular}
\end{table}

\section{Discussion}

In this work, we have presented a new model for dynamically recognizing the memory capacity of Hopfield networks. Our model runs alongside a learning Hopfield network and dynamically concludes at an accurate capacity estimate that minimizes memory loss and maximizes efficiency. Moreover, our model solves this problem for increasingly large networks and is moderately robust to variations in bias and correlation. We recognize a computational and memory tradeoff when using our model, as it has complexity $O(N^2)$ from calculating $W^{\mu-1}x^{\mu}$ and requires $N\times M$ memory to store crosstalk for all $M$ patterns in each $N$ neuron. However, these costs are on the same order as the $O(N^2)$ process for storing and recalling a pattern and the $N\times N$ memory needed to store the weight matrix. 

Our dynamic model enables enormous improvements on prior static methods by increasing the effective capacity of an $N$ neuron Hopfield network while ensuring near-perfect recall of all stored patterns. It also opens up possibilities for system-level improvements and algorithms such as reducing the computational overhead by reusing the pattern-recall pathway for calculating the new crosstalk, and selectively accepting or rejecting patterns based on their impact on the network's crosstalk. We feel these are powerful and interesting applications of our dynamic capacity estimation model. Our model can have significant implications for modern designers creating low-power, online learning systems, and repositions Hopfield networks as practical and efficient forms of associative memory.

\small

\bibliographystyle{plainnat}
\bibliography{HN_bibliography}

\end{document}